\documentclass[runningheads]{llncs}
\usepackage{graphicx}
\usepackage{adjustbox}
\usepackage{amssymb}
\usepackage{latexsym}
\usepackage{url}
\usepackage{xcolor}
\usepackage{amssymb,amsmath}
\usepackage{latexsym}
\usepackage{url}
\usepackage{xcolor}
\usepackage{algorithm}
\usepackage[noend]{algpseudocode}
\usepackage{bbm}
\usepackage{dsfont}
\usepackage{multirow}
\usepackage[percent]{overpic}
\usepackage{sidecap}

\def\unet{\mbox{U-Net}}
\def\eg{\emph{e.g.}}
\def\ie{{\it i.e.}}

\renewcommand{\Re}[1]{\mbox{$\mathbbm{R}^{#1}$}}

\begin{document}
\title{A Weakly Supervised Method for Instance Segmentation of Biological Cells}

\author{Fidel A. Guerrero-Pe\~na\inst{1,2}, Pedro D. Marrero Fernandez\inst{2}\\Tsang Ing Ren\inst{2}, Alexandre Cunha\inst{1}}

\authorrunning{Guerrero Pe\~na {\it et al.}}
\institute{Center for Advanced Methods in Biological Image Analysis -- CAMBIA\\California Institute of Technology, Pasadena CA, USA \and Center for Informatics, Federal University of Pernambuco, Recife PE, Brazil}

\maketitle              %
\section*{Abstract} 
We present a weakly supervised deep learning method to perform instance
segmentation of cells present in microscopy images.
Annotation of biomedical images in the lab can be scarce, incomplete, and
inaccurate. 
This is of concern when supervised learning is used for image analysis as the discriminative 
power of a learning model might be compromised in
these situations. To overcome the curse of poor labeling, our method focuses on three 
aspects to improve learning: 
i) we propose a loss function operating in three classes to facilitate separating adjacent cells and 
to drive the optimizer to properly classify underrepresented regions; 
ii) a contour-aware weight map model is introduced to strengthen contour detection 
while improving the network generalization capacity; and
iii) we augment data by carefully modulating local intensities on edges shared by adjoining regions
and to account for possibly weak signals on these edges.
Generated probability maps are segmented using different methods, with the
watershed based one generally offering the best solutions, specially in those
regions where the prevalence of a single class is not clear.
The combination of these contributions allows segmenting individual cells on
challenging images. We demonstrate our methods in sparse and crowded cell
images, showing improvements in the learning process for a fixed network
architecture.

\section{Introduction}
\begin{figure*}[t!]
  \begin{center}
    \includegraphics[width=0.495\linewidth]{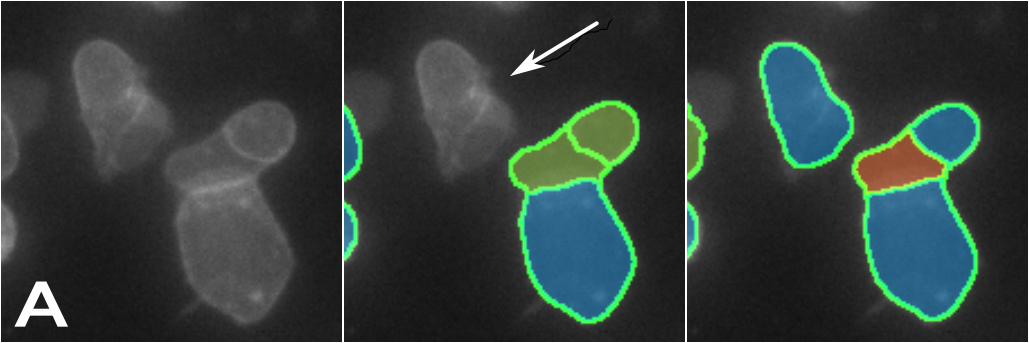}
    \hfill
    \includegraphics[width=0.495\linewidth]{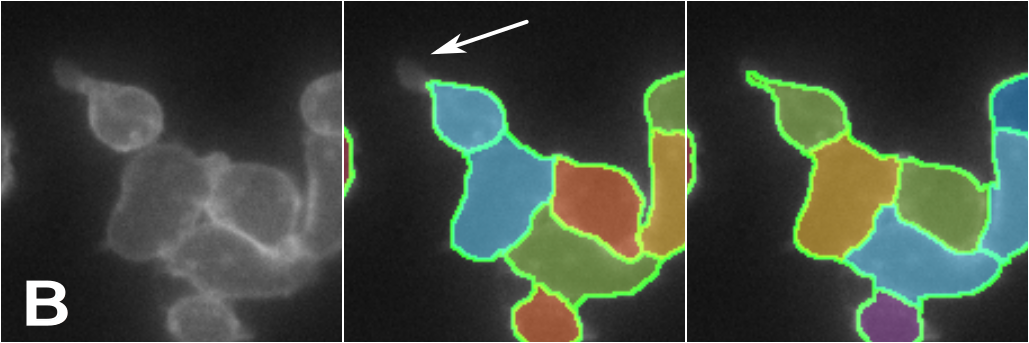}
    \vspace{-6.0mm}
    \caption{\fontsize{8}{8}
      Incomplete (A) and inaccurate (B) annotations of training images might be
      harmful for supervised learning as the presence of similar regions with
      erratic annotations might puzzle the optimization process.
      Our formulation is able to segment well under uncertainty as shown in the
      examples in the right panels of A and B above.
    }
    \vspace{-2.0mm}
    \noindent{\color{lightgray}\rule{1.0\columnwidth}{0.2mm}}
    \label{fig:res2}
  \end{center}\vspace{-1.6cm}
\end{figure*}

In developmental cell biology studies, one generally needs to quantify temporal
signals, \eg\ protein concentration, on a per cell basis.  This requires
segmenting individual cells in many images, accounting to hundreds or thousands
of cells per experiment. Such data availability suggests carrying on large
annotation efforts, following the common wisdom that massive annotations are
beneficial for fully supervised training to avoid overfitting and improve
generalization.
However, full annotation is expensive, time consuming, and it is often
inaccurate and incomplete when it is done at the lab, even by specialists (see
Fig.\ref{fig:res2}).

To mitigate these difficulties and make the most of limited training data, we
work on three fronts to improve learning. In addition to the usual data
augmentation strategies (rotation, cropping, etc.), we propose a new
augmentation scheme which modulates intensities on the borders of adjacent
cells as these are key regions when separating crowded cells. This scheme
augments the contrast patterns between edges and cell interiors. We also
explicitly endow the loss function to account for critically underrepresented
and reduced size regions so they can have a fair contribution to the functional
during optimization.  By adopting large weights on short edges separating
adjacent cells we increase the chances of detecting them as they now contribute
more significantly to the loss.  In our experience, without this construction,
these regions are poorly classified by the optimizer -- weights used in the
original \unet\ formulation \cite{ronneberger2015u} are not sufficient to
promote separation of adjoining regions. Further, adopting a three classes
approach \cite{guerrero2018multiclass} has significantly improved the
separation of adjacent cells which are otherwise consistently merged when
considering a binary foreground and background classification strategy. We have
noticed that complex shapes, \eg\ with small necks, slim invaginations and
protrusions, are more difficult to segment when compared to round, mostly
convex shapes \cite{schmidt2018cell}.  Small cells, tiny edges, and slim
parts, equally important for the segmentation result, can be easily dismissed
by the optimizer if their contribution is not explicitly accounted for and on
par with other more dominant regions.

{\bf Previous Work}. In \cite{kervadec2019constrained} the authors propose a weakly semantic
segmentation method for biomedical images. They include prior knowledge in the
form of constraints into the loss function for regularizing the size of
segmented objects. The work in \cite{yang2017suggestive} proposes a way to keep
annotations at a minimum while still capturing the essence of the signal
present in the images.  The goal is to avoid excessively annotating redundant
parts, present due to many repetitions of almost identical cells in the same
image. In \cite{liang2018weakly} the authors also craft a tuned loss function
applied to improve segmentation on weakly annotated gastric cancer images.
The instance segmentation method for natural images Mask R-CNN
\cite{he2017mask} uses two stacked networks, with detection followed by
segmentation. We use it for comparisons on our cell images.
Others have used three stacked networks for semantic segmentation and
regression of a watershed energy map allowing separating nearby objects
\cite{bai2017deep}. 
\vspace{-2mm}
\section{Segmentation Method}
\vspace{-2mm}

{\bf Notation}. 
Let $S=\{(x_j,g_j)\}_{j=1}^N$ be a training instance segmentation set
where $x_j\colon\Omega\to\Re{}^+$ is a single channel gray image defined on the
regular grid $\Omega\in\Re{2}$, and $g_j\colon\Omega\to\{0,\ldots,m_j\}$ its
instance segmentation ground truth map which assigns to a pixel $p\in\Omega$ a
unique label $g_j(p)$ among all $m_j + 1$ distinct instance labels, one for
each object, including background, labeled 0.
For a generic $(x,g)$, $V_i = \{p\; | \; g(p) = i\}$ contains all pixels
belonging to instance object $i$, hence forming the connected component
of object $i$. Due to label uniqueness, $V_i\cap V_j = \O, i\neq j$, \ie\
a pixel cannot belong to more than one instance thus satisfying the panoptic
segmentation criterion \cite{kirillov2018panoptic}.
Let $h\colon\Omega\to\{0,\ldots,C\}$ be a semantic segmentation map, obtained
using $g$, which reports the semantic class of a pixel among the $C+1$ possible
semantic classes, and $y\colon\Omega\to\Re{C+1}$ its one hot encoding
mapping. That is, for vector $y(p)\in\Re{C+1}$ and its $l$-th component
$y_l(p)$, we have $y_l(p)=1$ \text{iff} $h(p)=l$, otherwise $y_l(p)=0$. We call
$n_l=\sum_{p\in\Omega} y_l(p)$ the number of pixels of class $l$, and $\eta_k(p),
k\geqslant 1$, the $(2k+1)\times(2k+1)$ neighborhood of a pixel $p \in \Omega$. In
our experiments we adopted $k = 2$.

{\bf From Instance to Semantic Ground Truth}.
We formulate the instance segmentation problem as a semantic segmentation
problem where we obtain object segmentation and separation of cells at once.
To transform an instance ground truth to a semantic ground truth, we adopted
the three semantic classes scheme of \cite{guerrero2018multiclass}:
image background, cell interior, and touching region between cells.
This is suitable as the intensity distribution of our images in
those regions is multi-modal. We define our semantic ground truth $h$ as
\vspace{-2mm}
\begin{equation}\label{eq:gtoh}
    h(p) = \left\{ \begin{aligned}
    0 &\quad \text{ if } g(p)=0  &  \text{-- background} \\
    2 &\quad \text{ if } \sum\nolimits_{p^\prime\in\eta_k(p)}
    [g(p^\prime)\neq g(p)] \cdot [g(p^\prime)\neq 0] > 1 & \text{-- touching} \\
    1 &\quad \text{ otherwise} & \text{-- cell}
    \end{aligned} \right.
\vspace{-1mm}
\end{equation}
where $[\cdot]$ refers to Iverson bracket notation \cite{berman2018lovasz}:
$[b]=1$ if the boolean condition $b$ is true, otherwise $[b]=0$.
Eq.\ref{eq:gtoh} assigns class $0$ to all background pixels, it assigns class $2$
to all pixels whose neighborhood $\eta_k$ contains at least one pixel of another
connected component, and it assigns class $1$ to cell pixels not belonging to
touching regions.
\begin{figure*}[h!]
    \centering
    \setlength{\tabcolsep}{1pt}
    \begin{tabular}{ccccc}
    \begin{overpic}[width=0.195\textwidth]{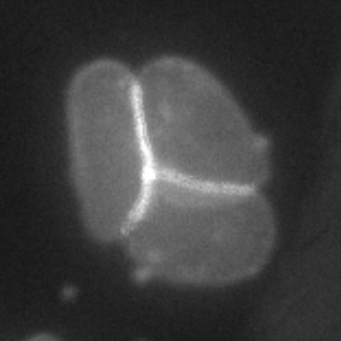}
    \put (65,85) {\color{white}$\displaystyle -1.0$}
    \end{overpic}&
    \begin{overpic}[width=0.195\textwidth]{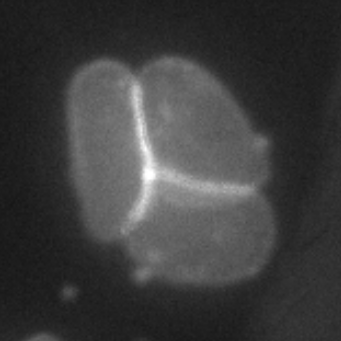}
    \put (65,85) {\color{white}$\displaystyle -0.5$}
    \end{overpic}&
    \begin{overpic}[width=0.195\textwidth]{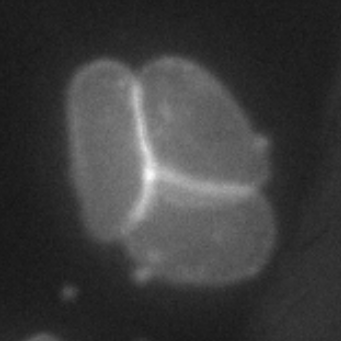}
    \put (70,85) {\color{white}$\displaystyle 0.0$}
    \end{overpic}&
    \begin{overpic}[width=0.195\textwidth]{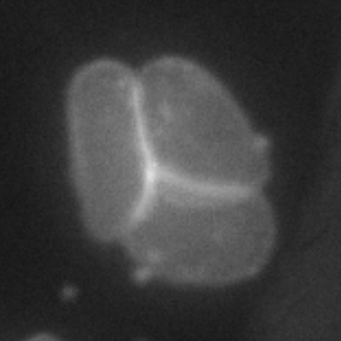}
    \put (70,85) {\color{white}$\displaystyle 0.5$}
    \end{overpic}&
    \begin{overpic}[width=0.195\textwidth]{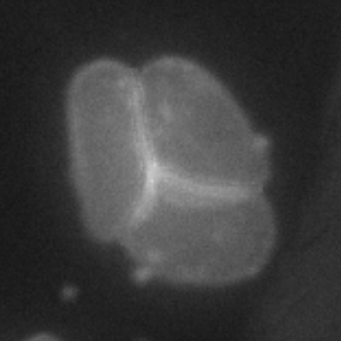}
    \put (70,85) {\color{white}$\displaystyle 1.0$}
    \end{overpic}\\
    \end{tabular}
    \vspace{-5mm}  
    \caption{{\footnotesize Contrast modulation around touching regions.
        Separating adjacent cells is one of the major challenges in crowded
        images.  To leverage learning, we feed the network with a variety of
        contrasts around touching regions. We do so by modulating their
        intesities while keeping adjacencies the same. In this example, an
        original image ($a =0$) has its contrast increased (decreased) around
        shared edges when we set $a < 0$ ($a > 0$).}}
    \label{fig:taug}
    \vspace{-2mm}
    \noindent{\color{lightgray}\rule{1.0\columnwidth}{0.2mm}}
    \vspace{-10mm}
\end{figure*}

{\bf Touching Region Augmentation}.  Touching regions have the lowest pixel
count among all semantic classes, having few examples to train the network.
They are in general brighter than their surroundings, but not always, with
varying values along its length. To train with a larger gamut of touching
patterns, including weak edges, we augment existing ones by modulating their
pixel values according to the expression $x_a(p) = (1-a)\cdot x(p)+a\cdot
\tilde{x}(p)$, only applied when $h(p) = 2$, where $\tilde{x}$ is the $7\times
7$ median filtered image of $x$.  When $a<0\; (a > 0)$ we increase (decrease)
contrast.  During training, we have random values of $a\sim U(-1,1)$.  An
example of this modulation is shown in Fig.\ref{fig:taug}.

{\bf Loss Function}.
\unet\ \cite{ronneberger2015u} is an encoder--decoder network for
biomedical image segmentation with proven results in small
datasets, and with cross entropy being the most commonly adopted loss function.
The weighted cross entropy \cite{ronneberger2015u} is a generalization 
where a pre--computed weight map assigns to each pixel its importance for the learning
process,
\vspace{-2mm}
\begin{equation}
    \mathcal{L}(y,z)= - \sum_{l=0}^{C} \sum_{p\in\Omega} \omega_{\beta,\nu,\sigma}(p) \cdot y_l(p) \cdot \log z_l(p)
\vspace{-1mm}
\end{equation}
where $\omega_{\beta,\nu,\sigma}(p)$ is the parameterized weight at pixel $p$, and $z_l(p)$ 
the computed probability of $p$ belonging to class $l$ for ground truth $y_l(p)$.

Let $R(u)=u^+$ be the rectified linear function, ReLu, and $\varphi_{\beta}(u)=
R(1- u/\beta), u\in\Re{}$, a rectified inverse function saturated in $\beta\in\Re{}^+$. We
propose the Triplex Weight Map, $W^3$, model
\vspace{-2mm}
\begin{equation}\label{eq:weight}
    \omega_{\beta,\nu,\sigma}(p)=\begin{cases}
    \nu/n_0+\nu\cdot \varphi_{\beta}\left(\phi_{h}(p)\right)/n_1 & \text{ if } h(p)=0 \\
    \nu/n_1 + \nu\cdot  \varphi_{\beta}\left(\phi_K(p)\right) & \text{ if }  h(p)=1, p \in \Gamma \\
    \nu/n_1 +  \omega_{\beta,\nu,\sigma}(\zeta_{\Gamma}( p)) \cdot  \exp(-\phi_{\Gamma}^2(p)/\sigma^2) & \text{ if }  h(p)=1, p\notin \Gamma \\
     \nu/n_2 & \text{ if } h(p)=2 \\
    \end{cases}
\vspace{-1mm}
\end{equation}
where $\Gamma$ represents cell contour; $n_l$ is the number of pixels of class
$l$; $\phi_{h}$ is the distance transform over $h$ that assigns to every pixel
its Euclidean distance to the closest non-background pixel; $\phi_K$ and
$\phi_{\Gamma}$ are, respectively, the distance transforms with respect to the
skeleton of cells and cell contours; and $\zeta_{\Gamma}\colon\Omega\to\Omega$
returns the pixel in contour $\Gamma$ closest to a given pixel $p$, thus
$\zeta_{\Gamma}(p)\in\Gamma$. The $W^3$ model sets
$\omega_{\beta,\nu,\sigma}(p) = \nu/n_0$ for all background pixels distant at
least $\beta$ to a cell contour.
This way, true cells that are eventually not annotated and located beyond $\beta$ from
annotated cells have very low importance during training -- by design, weights on
non annotated regions are close to zero.

The recursive expression for foreground pixels (third line in eq.3) creates
weights using a rolling Gaussian with variance $\sigma^2$ centered on each
pixel of the contour. These weights have amplitudes which are inversely
proportional to their distances to cell skeleton, resulting in large values
for slim and neck regions. The parameter $\nu$ is used for setting the
amplitude of the Gaussians. The weight at a foreground pixel is the value of
the Gaussian at the contour point closest to this pixel.
The
touching region is assigned a constant weight for class balance, larger than all other weights.

{\bf From Semantic to Instance Segmentation}.
After training the network for semantic segmentation, we perform the
transformation from semantic to panoptic, instance segmentation. First, a
decision rule $\hat{h}$ over the output probability map $z$ is applied to hard
classify each pixel.  The usual approach is to classify with {\it maximum a
posteriori} (MAP) where the semantic segmentation is obtained with
$\hat{h}(p)=\arg\max_l z_l(p)$.
However, since pixels in the touching and interior cell regions share similar
intensity distributions, the classifier might be uncertain in the transition
zone between these regions, where it might fail to assign the right class
for some, sometimes crucial, pixels in these areas.  A few misclassified pixels
can compromise the separation of adjacent cells (see Fig.\ref{fig:map}).
Therefore, we cannot solely rely on MAP as our hard classifier.
An alternative is to use a thresholding (TH) strategy as a decision rule,
where parameters $\gamma_1$ and $\gamma_2$ control,
respectively, the class assignment of pixels:
$\hat{h}(p)= 2 \text{ if } z_2(p)\geq\gamma_2$, and 
$\hat{h}(p)= 1 \text{ if } z_1(p)\geq\gamma_1 \text{ and } z_2(p)<\gamma_2$, and $0$ otherwise.
Finally, the estimated instance segmentation $\hat{g}$ labels each cell region
$\hat{V}_i$ and it distributes touching pixels to their closest components,
\vspace{-1mm}
\begin{equation}\label{eq:th}
    \hat{g}(p)=\begin{cases}
        0 & \text{ if }\hat{h}(p)=0\\
        i & \text{ if }\hat{h}(p)=1 \textbf{ and } p \in \hat{V}_i\\
        \hat{g}(\zeta_{\Gamma}(p)) & \text{ if }\hat{h}(p)=2
    \end{cases}
\vspace{-1mm}
\end{equation}
Another alternative for post-processing is to segment using the Watershed
Transform (WT) with markers. It is applied on the topographic map formed by the
subtraction of touching and cell probability maps, $z_2 - z_1$. Markers are
comprised of pixels in the background and cell regions whose probabilities are
larger than given thresholds $\tau_0$ and $\tau_1$, $\{ p | z_0(p)\geq\tau_0
\text{ or } z_1(p)\geq\tau_1 \}$. High values for these should be safe, \eg\
$\tau_0 = \tau_1 = 0.8$.
\vspace{-2mm}
\section{Experiments and Results}
\label{sec:results}
\vspace{-2mm}

Training of our triplex weight map method, $W^3$, is done using U-Net
\cite{ronneberger2015u} initialized with normally distributed weights according
to the Xavier method \cite{glorot2010understanding}.  We compare it to the
following methods: Lov{\'a}sz-Softmax loss function ignoring the background
class, LSMAX \cite{berman2018lovasz}; weighted cross entropy using class
balance weight map, BWM; U-Net with near object weights \cite{ronneberger2015u}
adapted to three classes, UNET; and the per-class average combination of the
probability maps from BWM, UNET, and $W^3$, followed by a softmax, named COMB.
We also compared our results with those obtained by Mask R-CNN, MRCNN
\cite{he2017mask}. The use of COMB is motivated by ensenble classifiers where
one tries to combine the predictions of multiple classifiers to achieve a
prediction which is potentially better than each individual one.  We plan to
explore other choices beyond averaging.

We trained all networks over a cell segmentation dataset containing 28 images
of size 1024x1024 with weak supervision in the form of incomplete and
inaccurate annotations. We use the optimizer Adam with initial learning rate
$lr=10^{-4}$.  The number of epochs and minibatch size were, respectively, $1000$
and $1$.  We augmented data during training by random mirroring,
rotating, warping, gamma correction, and touching contrast modulation, as in
Fig.2.

We follow \cite{kirillov2018panoptic} to assess results. For detection, we use
the Precision (P05) and the Recognition Quality (RQ) of instances with Jaccard
index above 0.5. For segmentation, we use Segmentation Quality
(SQ) computed as the average Jaccard of matched segments.
For an overall evaluation of both detection and
segmentation, we use the Panoptic Quality (PQ)
metric, $PQ=RQ\cdot SQ$.

\begin{figure*}[t!]
    \centering
    \setlength{\tabcolsep}{2pt}
    \begin{tabular}{cccccc}
    \includegraphics[width=.165\linewidth]{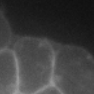}&
    \includegraphics[width=.165\linewidth]{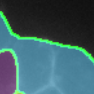}&
    \includegraphics[width=.165\linewidth]{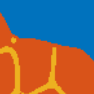}&
    \includegraphics[width=.165\linewidth]{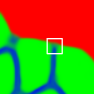}&
    \includegraphics[height=.165\linewidth]{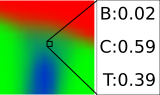}\\
    \scriptsize{Image} & \scriptsize{Segmentation} & \scriptsize{MAP} & \scriptsize{Prob. map} & \scriptsize{Prob. values} \\
    \end{tabular}
    \vspace{-4mm}
    \caption{{\small Poor classification. 
        {\it Maximum a posteriori}, MAP, does not separate adjacent cells due to 
        poor probabilities in the junctions shown above. The misclassification of just a few pixels renders a wrong cell
        topology. Probability maps are shown as RGB images with 
        Background (red), Cell (green) and Touching (blue) classes.}}
    \label{fig:map}
    \vspace{-2mm}
    \noindent{\color{lightgray}\rule{1.0\columnwidth}{0.2mm}}
    \vspace{-10mm}
\end{figure*}

\begin{figure*}[b!]
    \footnotesize
    \setlength{\tabcolsep}{0pt}
    \begin{tabular}{ccc}
    \begin{overpic}[width=0.33\textwidth]{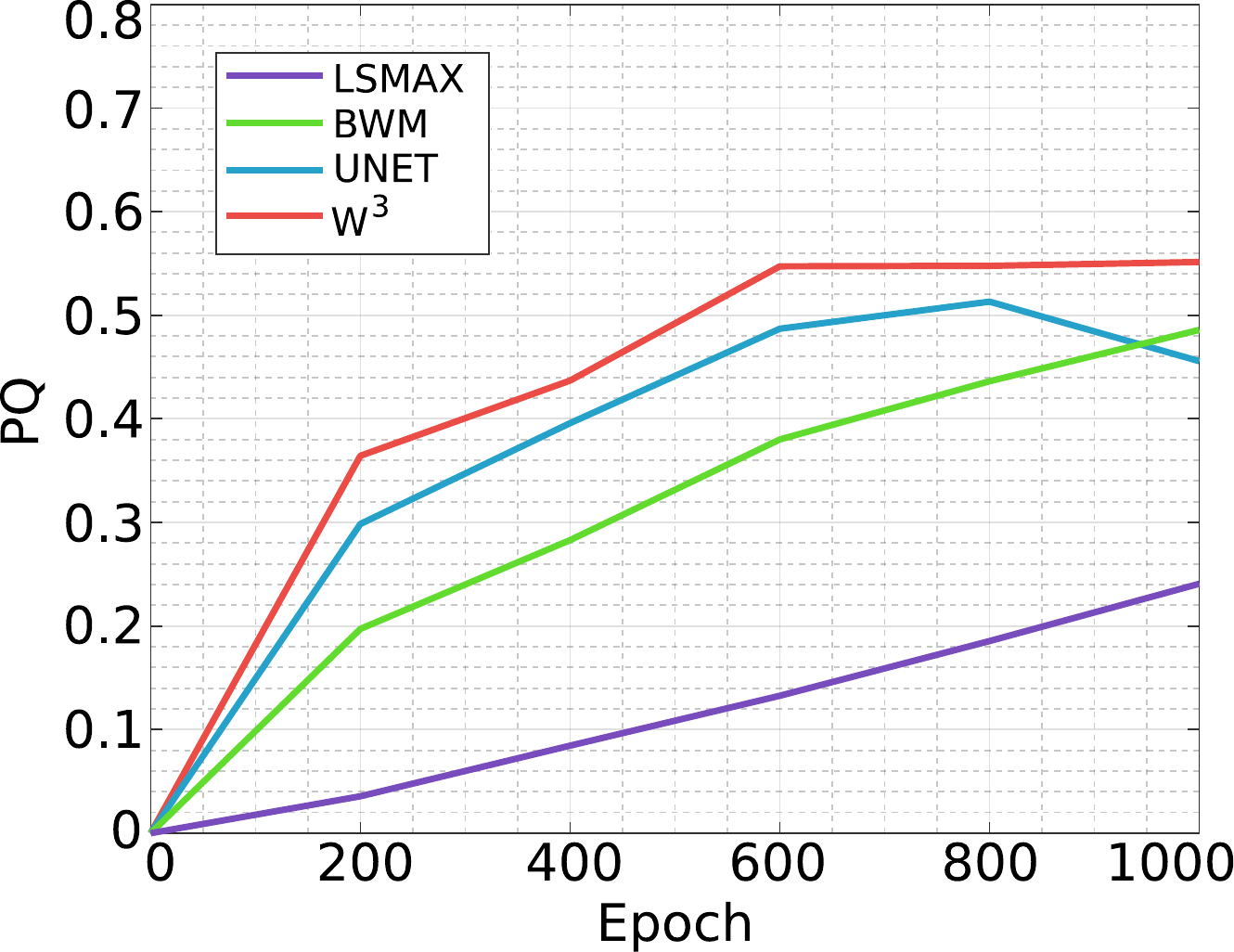}
    \put (14,78) {{\small Maximum A Posterior}}
    \end{overpic}&
    \begin{overpic}[width=0.33\textwidth]{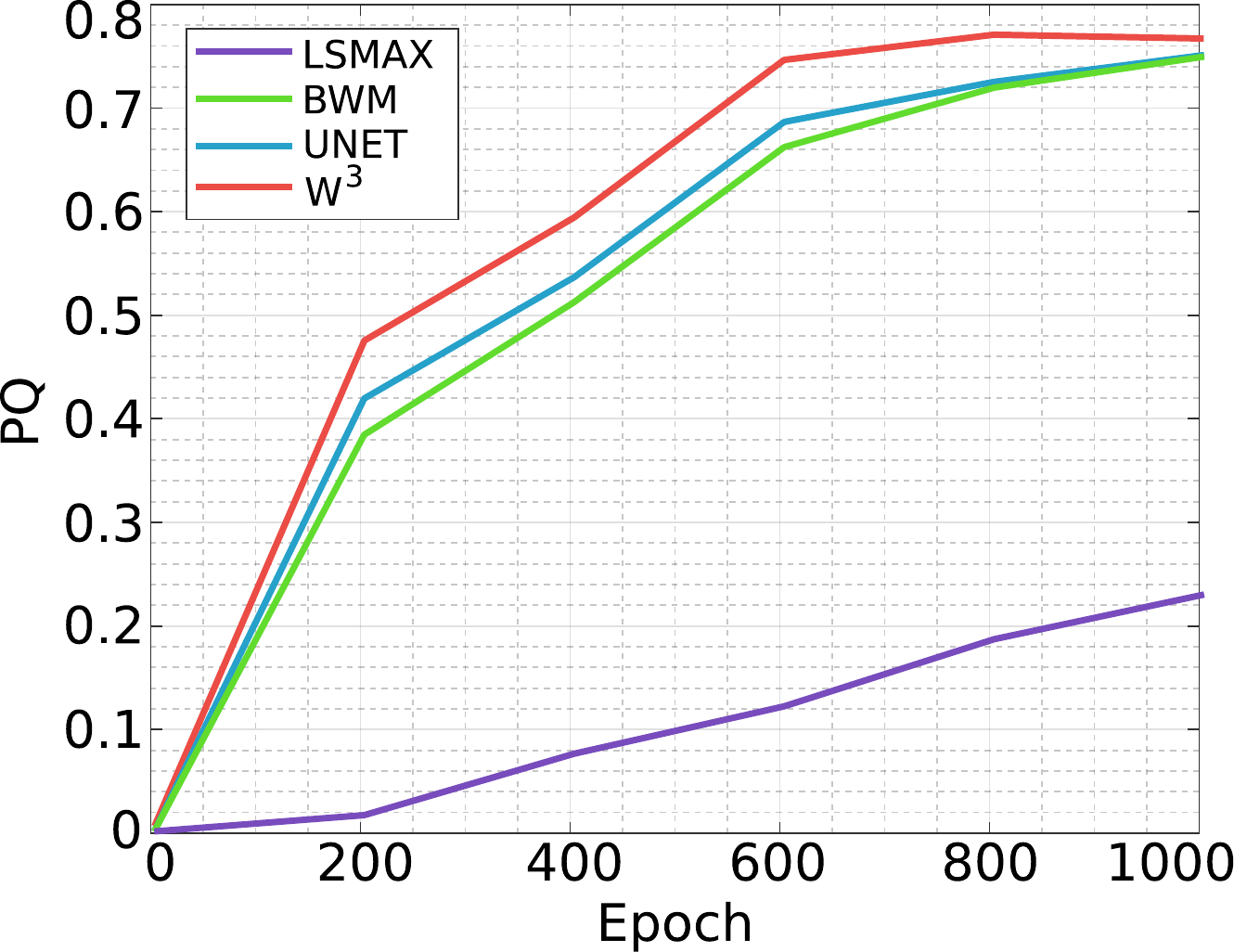}
    \put (35,78) {{\small Threshold}}
    \end{overpic}&
    \begin{overpic}[width=0.33\textwidth]{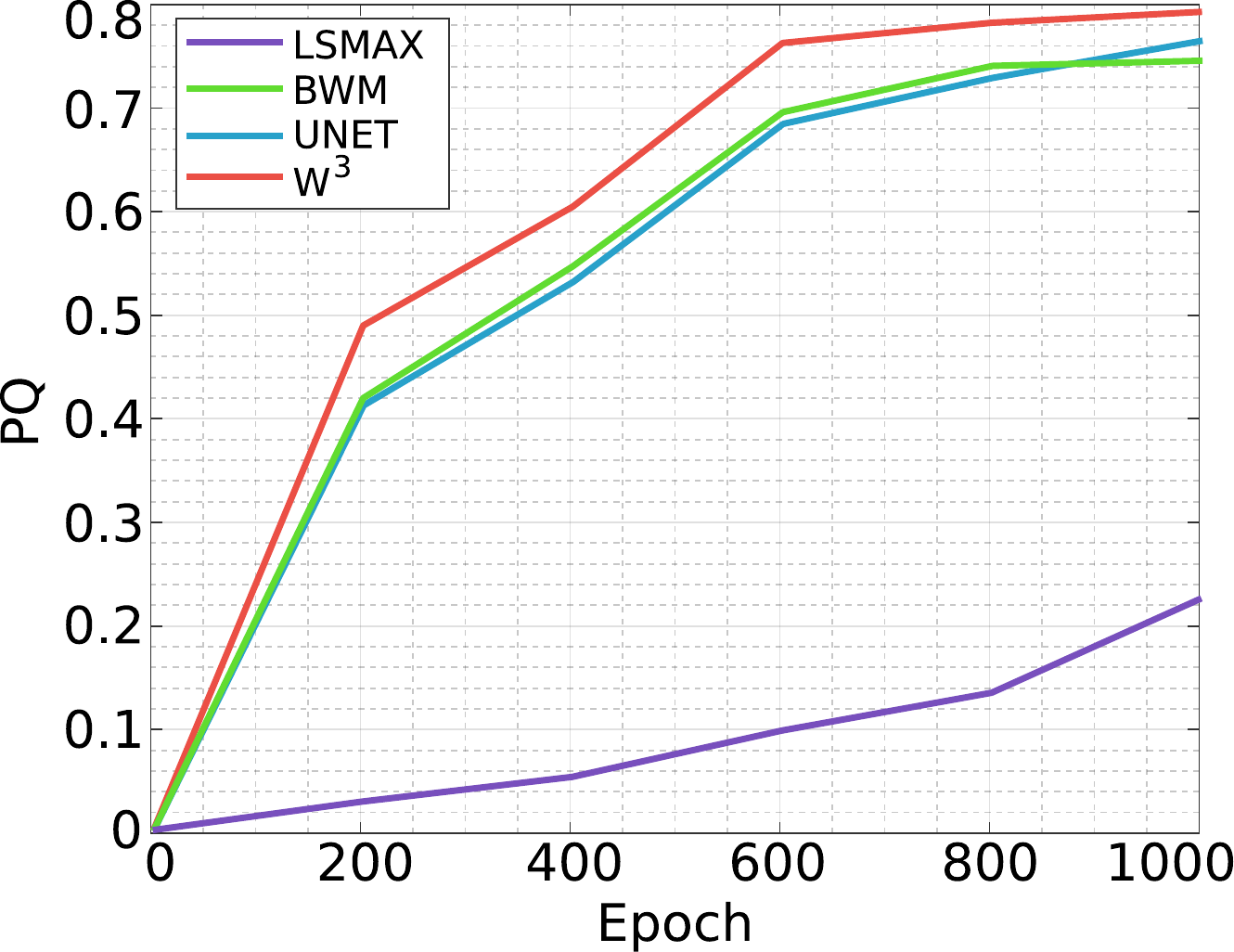}
    \put (35,78) {{\small Watershed}}
    \end{overpic}\\
    \end{tabular}
    \vspace{-4mm}
    \caption{{\small Panoptic Quality (PQ) training values for all methods we
        compare to $W^3$, except COMB, 
        using {\it Maximum a Posteriori}
        (MAP), Thresholded Maps (TH) and Watershed Transform (WT)
        post-processing. $W^3$ converges faster to a better solution.}}
    \label{fig:pqtime}
\end{figure*}
{\bf Panoptic Segmentation Performance}.  
We performed an exploration over the parameter space for the two parameters
used in the TH and WT postprocessing methods.  Table \ref{tab:posproc} shows a
comparison of different post-processing strategies considering the best
combination of parameters for Thresholds (TH) and Watershed (WT).
For Mask R-CNN we used the same single threshold TH on the instance probability maps of
all boxed cells. We performed watershed WT on each boxed cell region with seeds
extracted from the most prominent background and foreground regions in the
probability maps.
Although Lov{\'a}sz-Softmax seems to be a promising loss function, we believe
that the small training dataset and minibatch size negatively influenced its
performance.  For most values of thresholds used in the TH post-processing, the
average combination (COMB) improved the overall result due to the reduction of
False Positives (see P05 column). Also, in most cases, our $W^3$ approach
obtained better SQ values than other methods suggesting a better contour
adequacy.  Because touching and cell intensity distributions overlap, a softer
classification was obtained for these regions. MAP did not achieve the same
performance of other approaches (Fig. \ref{fig:map}). The behavior in Table
\ref{tab:posproc} remained the same during training as shown in Fig.
\ref{fig:pqtime}.
\begin{table}[b!]
\noindent{\color{lightgray}\rule{1.0\columnwidth}{0.2mm}}
\caption{{\small
    Metric values for different post processing schemes and segmentation methods. Numbers are average values obtained for the best combination of threshold parameters for both TH and WT post processing methods. Tests were done on 7 images, totaling 138 cells, with 14 clusters containing from 2 to approximately 32 cells. Metric values obtained with TH and WT are higher than those obtained with MAP showing that our post procesing schemes improve results. 
Overall, our $W^3$ and COMB outperform other segmentation methods for almost all metrics, except P05, when thresholding and watershed classification schemes are adopted.
}}
\label{tab:posproc}
\vspace{-3mm}
\resizebox{\columnwidth}{!}{%
\begin{tabular}{|l|c|c|c|c|c|c|c|c|c|c|c|c|} 
\hline
\multirow{2}{*}{Methods} & \multicolumn{4}{c|}{MAP} & \multicolumn{4}{c|}{TH} & \multicolumn{4}{c|}{WT}  \\ 
\cline{2-13}
        & P05  & RQ    & SQ    & PQ    & P05  & RQ    & SQ    & PQ    & P05  & RQ    & SQ    & PQ    \\ 
\hline
MRCNN    &\textbf{0.9188} &\textbf{0.8617} &0.8002 &\textbf{0.6892} &  \textbf{0.9343}    &  0.8767    &  0.8012   &  0.7019   &  \textbf{0.9343}   &   0.8767   &   0.8019   &  0.7026    \\
LSMAX    &0.3871 &0.3236 &0.7455 &0.2408 &0.4348 &0.3119 &0.7171 &0.2286 &0.4000 &0.3149 &0.7073 &0.2237 \\
BWM      &0.6756 &0.5580 &0.8674 &0.4858 &0.8583 &0.8504 &0.8769 &0.7476 &0.8193 &0.8405 &0.8831 &0.7437 \\
U-Net     &0.6801 &0.5381 &0.8418 &0.4556 &0.8413 &0.8508 &0.8791 &0.7492 &0.8708 &0.8600 &0.8850 &0.7621 \\
$W^3$(Ours)&0.7384 &0.6305 &\textbf{0.8721} &0.5513 &0.8477 &0.8439 &\textbf{0.8994} &0.7604 &0.9028 &\textbf{0.8775} &\textbf{0.8995} &\textbf{0.7896} \\
COMB(Ours)     &0.7587 &0.6129 &0.8698 &0.5351 &0.8952 &\textbf{0.8851} &0.8908 &\textbf{0.7889} &0.8925 &0.8759 &0.8944 &0.7837 \\
\hline
\end{tabular}%
}
\end{table}

\begin{figure*}[t!]
    \footnotesize
    \setlength{\tabcolsep}{0pt}
    \begin{tabular}{cccccccc}
    \includegraphics[width=.125\linewidth]{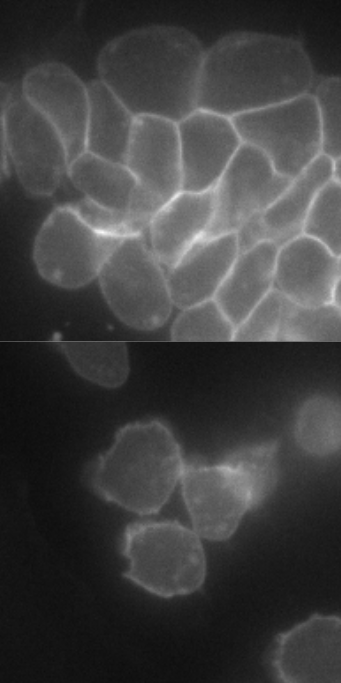}&
    \includegraphics[width=.125\linewidth]{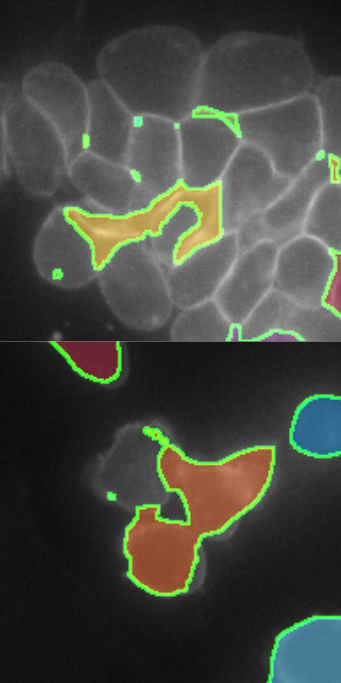}&
    \includegraphics[width=.125\linewidth]{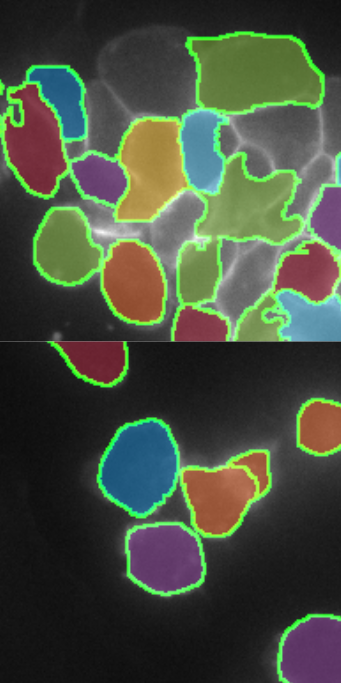}&
    \includegraphics[width=.125\linewidth]{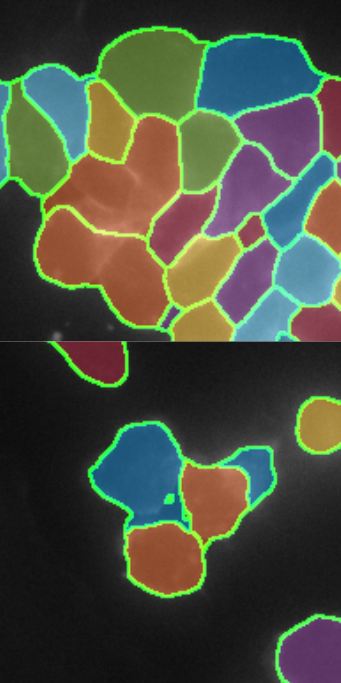}&
    \includegraphics[width=.125\linewidth]{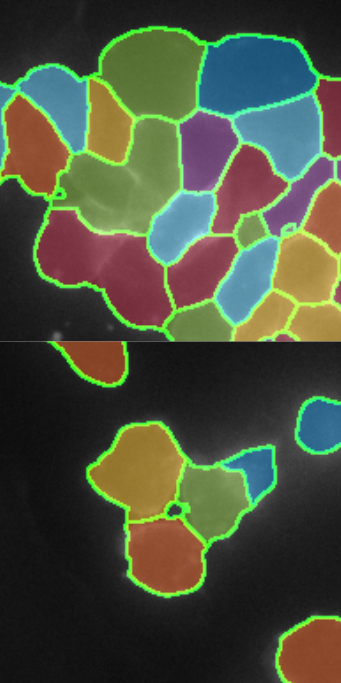}&
    \includegraphics[width=.125\linewidth]{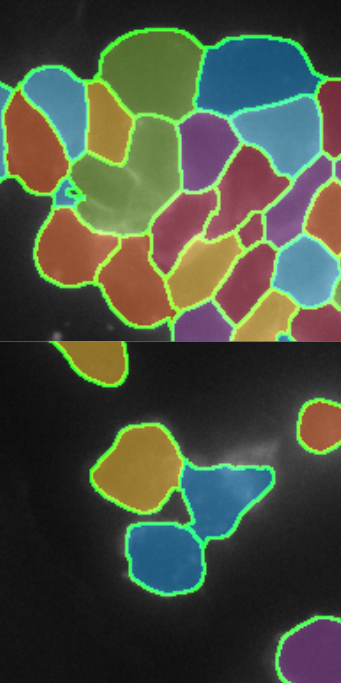}&
    \includegraphics[width=.125\linewidth]{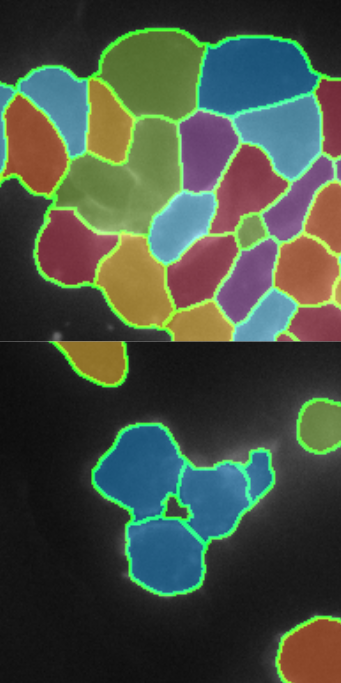}&
    \includegraphics[width=.125\linewidth]{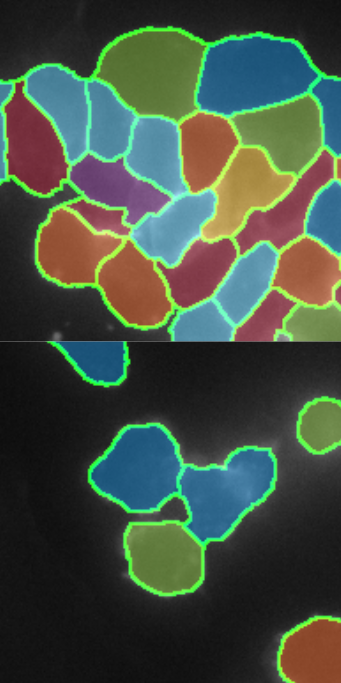}\\
    \scriptsize{Image} & \scriptsize{LSMAX} & \scriptsize{MRCNN} & \scriptsize{U-Net} & \scriptsize{BWM} & 
    \scriptsize{$W^3$} & \scriptsize{COMB} & \scriptsize{Annotation} \\
    \end{tabular}
    \vspace{-4mm}
    \caption{{\small Segmentation results for packed cell clusters obtained
        using methods described in section \ref{sec:results}.  Colors serve
        to show cell separation. Note the superiority of our $W^3$.
      }}
    \label{fig:res}
    \vspace{-2mm}
    \noindent{\color{lightgray}\rule{1.0\columnwidth}{0.2mm}}
    \vspace{-10mm}
\end{figure*}
\begin{figure*}[b!]
    \centering
    \setlength{\tabcolsep}{1pt}
    \begin{tabular}{cccc}
    \begin{overpic}[width=0.24\textwidth]{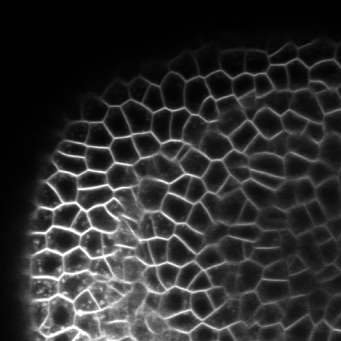}
    \put (25,88) {\color{white}Meristem}
    \end{overpic}&
    \begin{overpic}[width=0.24\textwidth]{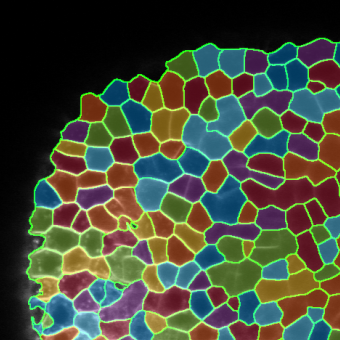}
    \put (15,88) {\color{white}Segmentation}
    \end{overpic}&
    \begin{overpic}[width=0.24\textwidth]{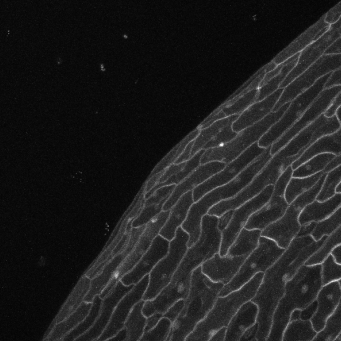}
    \put (35,88) {\color{white}Sepal}
    \end{overpic}&
    \begin{overpic}[width=0.24\textwidth]{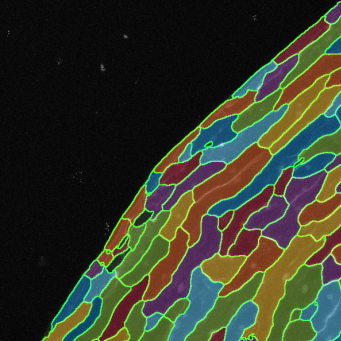}
    \put (15,88) {\color{white}Segmentation}
    \end{overpic}\\
    \end{tabular}
    \vspace{-4mm}
    \caption{{\small Zero-shot panoptic segmentation of meristem and sepal
        images with our $W^3$ method exclusively trained with cell images from different domains.
    }}
    \label{fig:res3}
\end{figure*}

Examples of segmenting crowded cells with various methods are shown in Fig. \ref{fig:res}. 
In our experiments, MRCNN was able to correctly segment isolated and nearly
adjacent cells (second row), but it sometimes failed in challenging
high-density clusters.  BWM and U-Net tend to misclassify background pixels in
neighboring cells (second row) with estimated contours generally beyond cell
boundaries.  $W^3$ had a better detection and segmentation performance with
improvement of contour adequacy over COMB.

We believe our combined efforts of data augmentation, loss formulation with per
pixel geometric weights, and multiclass classification enabled our trained
neural networks to correctly segment cells even from domains it has never seen.  For
example, we have never trained with images of meristem and sepal cells but we
still obtain good quality cell segmentation for these as shown in
Fig.\ref{fig:res3}.  These solutions might be further improved by training with
a few samples from these domains.

\vspace{-4mm}
\section{Conclusions}
\vspace{-2mm} We proposed a weakly supervised extension to the weighted cross
entropy loss function that enabled us to effectively segment
crowded cells. We used a semantic approach to solve a panoptic segmentation
task with a small training dataset of highly cluttered cells which have
incomplete and inaccurate annotations. A new contrast modulation was proposed
as data augmentation for touching regions allowing us to perform an adequate
panoptic segmentation. We were able to segment images from domains
other than the one used for training the network. The experiments showed
a better detection and contour adequacy of our method and a faster convergence
when compared to similar approaches. 

\bibliographystyle{splncs04}

\end{document}